\documentclass[letterpaper, 10 pt, conference]{ieeeconf}  

\IEEEoverridecommandlockouts
\let\labelindent\relax
\usepackage{enumitem}
\usepackage{graphicx}
\usepackage{subfig}
\usepackage{amsmath}
\usepackage{amssymb}
\usepackage{float}
\usepackage{color}
\usepackage[ruled,vlined,linesnumbered]{algorithm2e}
\usepackage{wrapfig}

\IEEEoverridecommandlockouts                              

\title{\LARGE \bf
InsertionNet 2.0: Minimal Contact Multi-Step Insertion Using Multimodal Multiview Sensory Input
}
\author{Oren Spector$^{1}$, Vladimir Tchuiev$^{1}$ and Dotan Di~Castro$^{1}$
\thanks{$^{1}$Bosch Center for Artificial Inteligence, Haifa, Israel. 
        {\tt\small oren.spector@il.bosch.com}}%
}

\definecolor{vova}{rgb}{0.05,0.5,0.05}

\begin{document}


\maketitle
\thispagestyle{empty}
\pagestyle{empty}

\begin{abstract}


We address the problem of devising the means for a robot to rapidly and safely learn insertion skills with just a few human interventions and without hand-crafted rewards or demonstrations. Our InsertionNet version 2.0 provides an improved technique to robustly cope with a wide range of use-cases featuring different shapes, colors, initial poses, etc. In particular, we present a regression-based method based on multimodal input from stereo perception and force, augmented with contrastive learning for the efficient learning of valuable features.
In addition, we introduce a one-shot learning technique for insertion, which relies on a relation network scheme to better exploit the collected data and to support multi-step insertion tasks. 
Our method improves on the results obtained with the original InsertionNet, achieving an almost perfect score (above 97.5$\%$ on 200 trials) in 16 real-life insertion tasks while minimizing the execution time and contact during insertion. We further demonstrate our method's ability to tackle a real-life 3-step insertion task and perfectly solve an unseen insertion task without learning.

\end{abstract}


\section{INTRODUCTION} \label{sec:introduction}

Constructing a robot that reliably inserts diverse objects (e.g., plugs, engine gears) is a grand challenge in the design of manufacturing, inspection, and home-service robots. Minimizing action time, maximizing reliability, and minimizing contact between the grasped object and the target component is difficult due to the inherent uncertainty concerning sensing, control, sensitivity to applied forces, and occlusions. 

Previous works tackled this problem using Reinforcement Learning (RL), a machine learning approach known for its versatility and generality. RL methods involve executing learning algorithms such as off-policy RL  \cite{vecerik2019practical}, \cite{lee2020guided} and model-based RL \cite{luo2018deep} directly on the robot. While demonstrating generalization properties to some extent \cite{lee2019making}, this type of online interaction is impractical on contact-rich tasks, either because collecting data on a real robot is extremely expensive or because unpredicted moves created by initial policies can endanger the robot and its equipment (e.g., force torque sensor).


In this work, we introduce a data-efficient, safe, supervised approach to acquire a robot policy: InsertionNet 2.0, a substantial extension of previous versions of InsertionNet \cite{spector2021insertionnet} that learns minimal contact policies as well as multi-step insertion tasks with only a few data points by utilizing contrastive methodologies and one-shot learning techniques.

Our primary contributions and improvements over InsertionNet are as follows:
\begin{enumerate}[leftmargin=*]
	\item Using two cameras in order to avoid the one image ambiguity problem and extract depth information.
	\item Eliminating the requirement to touch the socket's surface when inserting an object.
	\item Including a relation network that enables one-shot learning and multi-step insertion.
	\item Integrating contrastive learning in order to reduce the amount of labeled data.
\end{enumerate}

\begin{figure}[t]
\centering
\includegraphics[width=0.3\textwidth]{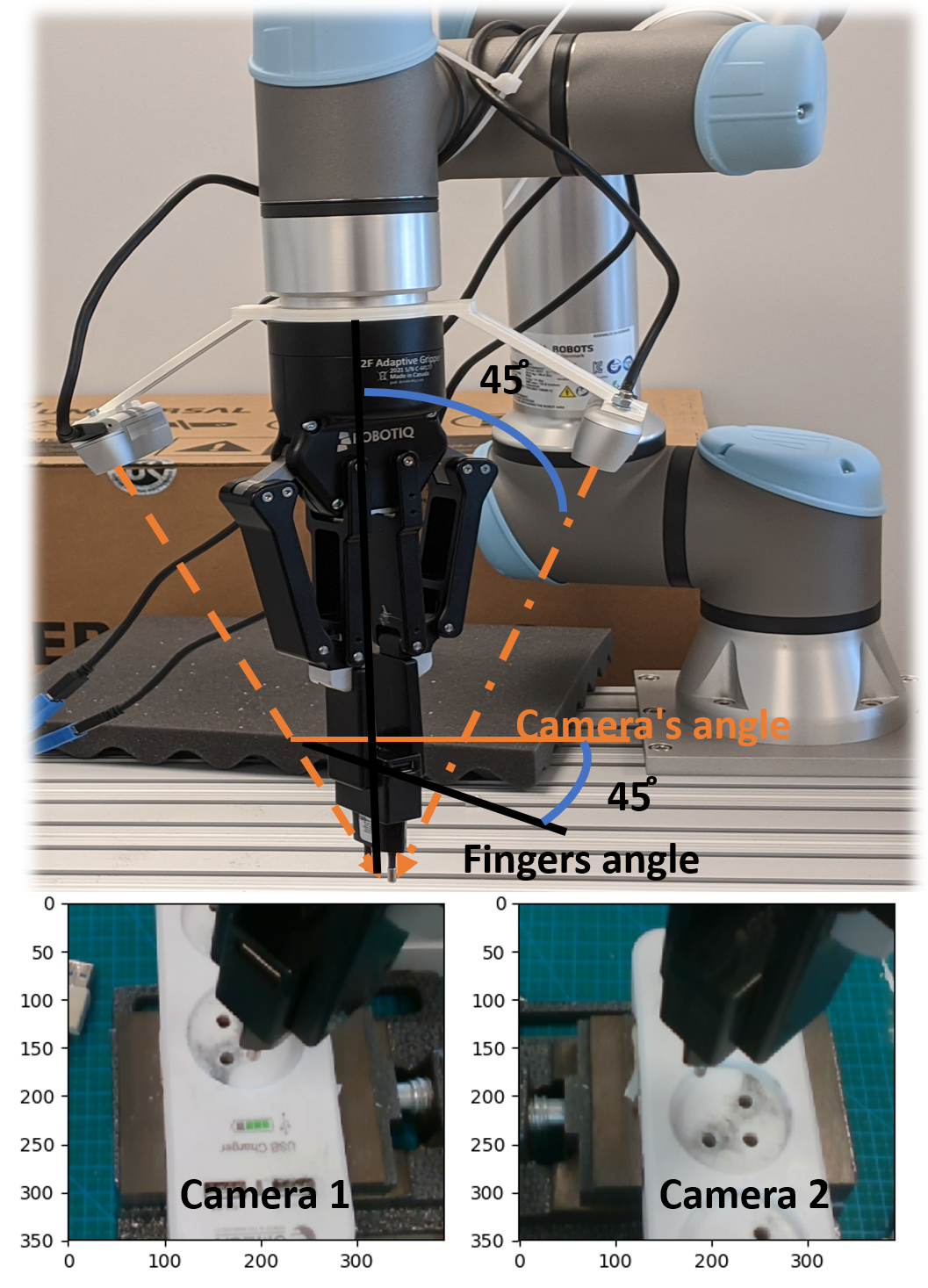}
\caption{\label{fig:system comp} The system's components include two RealSense D435 wrist cameras, tilted at a $45^\circ$ angle in relation to the gripper's axis, and focused on a point between the EFF fingers.}
\end{figure}


\section{Related work}
\subsection{Contact-rich Manipulation}

RL is a common approach in learning an adaptable and flexible solution for robotic manipulation \cite{kober2009learning} and insertion \cite{Xu2019,lee2019making}. In order to minimize the interaction with the environment during training (avoiding expensive time on the robot), current solutions utilize low sample complexity RL algorithms such as off-policy  \cite{vecerik2019practical,lee2020guided} and model-based \cite{luo2018deep} approaches. In addition, current solutions exploit prior knowledge on the tasks via a residual policy or the Learning from Demonstration (LfD) methodology, which have been shown to drastically shrink the problem sample's complexity \cite{schoettler2020deep} and to facilitate generalization. Residual policy methods \cite{johannink2019residual,lee2020guided} use a pre-program fixed policy while employing RL only to learn a correction policy. LfD is used to derive a baseline policy \cite{davchev2020residual,vecerik2019practical} or to initiate the learning process. While the above-noted approaches have shown significant results in very challenging problems, they are limited by the online RL scheme's ability to collect a large variety of data on the robot and utilize a large collected experience.

An alternative approach for enabling deep-RL in robots is Sim2Real: learning a policy in simulation and transferring it to a real robot. Such an approach has shown promising results, most notably, in grasping \cite{james2019sim} and, most recently, insertion tasks \cite{spector2020deep}. Sim2Real highlights the generalization ability of RL approaches in insertion tasks if a large amount of diverse data can be collected. However, this approach is limited by the domain expert's ability to model the real-world scenario in the simulation.


\subsection{Contrastive Learning}

Contrastive Learning is a framework of learning representations that obey similarity or dissimilarity constraints in a dataset that map onto positive or negative labels, respectively. A simple contrastive learning approach is Instance Discrimination (Wu et al., 2018), where an example and an image are a positive pair if they are data-augmentations of the same instance and negative otherwise. A key challenge in contrastive learning is the choice of negative samples, as it may influence the quality of the underlying representations learned. Recent works that employed contrastive learning in RL schemes have shown that adding a contrastive loss in complex vision-based control tasks may decrease the number of samples needed to learn an optimal policy \cite{srinivas2020curl}.

\subsection{One-Shot Learning}
Data augmentation and regularization techniques can alleviate overfitting in a limited-data regime, but they do not solve it. Using an embedding model is a common approach to tackle few-shot or small sample learning. This entails using the similarity function $\textrm{sim}(f(x), f(y))$, which measures the likeness between two embedded images, $f(x)$ and $f(y)$ \cite{wang2020generalizing}. The first such model
embedded samples using a kernel. Recently, more sophisticated embeddings have been trained with convolutional Siamese networks \cite{koch2015siamese}.
In addition to the embedding function, $f(\cdot)$, as recently shown \cite{sung2018learning}, a similarity function that can be learnt outperforms a fixed nearest-neighbour or linear classifier \cite{koch2015siamese}. 

\section{METHOD} \label{sec:method}
In this section, we describe the system, data collection approach and methodology we use to train the residual policy network. 


\newcommand{\DX}{\Delta x}
\newcommand{\DY}{\Delta y}
\newcommand{\DZ}{\Delta z}
\newcommand{\DTX}{\Delta \theta_x}
\newcommand{\DTY}{\Delta \theta_y}
\newcommand{\DTZ}{\Delta \theta_z}
\newcommand{\DXd}{\Delta x^d}
\newcommand{\DYd}{\Delta y^d}
\newcommand{\DZd}{\Delta z^d}
\newcommand{\DTXd}{\Delta \theta^d_x}
\newcommand{\DTYd}{\Delta \theta^d_y}
\newcommand{\DTZd}{\Delta \theta^d_z}
\newcommand{\DELTA}{\Delta}
\newcommand{\IMG}{\textrm{Img}}

\subsection{The System's Components, Setup and Definitions}
\label{subsec:system_components}
We used a Universal Robot \emph{UR5e} arm (Figure \ref{fig:system comp}) with 6DoF. This robot has $2$ sensory inputs. The first is stereoscopic perception, endowed by two RealSense D435 wrist cameras tilted at a $45^\circ$ angle and focused on a point between the end effector fingers (EEF). Assuming a height $H$, width $W$ and $3$ channels for an image, we denote the image by $\IMG \in \mathbb{R}^{H \times W \times 6}$. The second input is force sensing on the EEF\footnote{using a force sensor calculated by the UR5e robotic arm.}, composed of force $F \triangleq (f_x, f_y, f_z)\in \mathbb{R}^3$ and moment $M=(m_x, m_y, m_z) \in \mathbb{R}^3$. We define the robot's observation as $O=(\IMG,F,M)$. To accurately capture the contact forces and generate smooth movements, we used high-frequency communication (Real Time Data
Exchange). Specifically, F/T and position measurements were sampled at 500Hz and effort commands were sent at 125Hz.

The arm has 6DoF and a gripper \cite{lynch2017modern}. We denote by $L$ the pose of the gripper. Specifically, $L\triangleq[L_\textrm{pos},L_\textrm{ang}]$, where $L_\textrm{pos}\triangleq(x_0, y_0, z_0)$ is the gripper's location, and $L_\textrm{ang}\triangleq(\theta^x_0, \theta^y_0, \theta^z_0)$ its pose. We define the robot's action in Cartesian space by $\DELTA=(\DX,\DY,\DZ,\DTX,\DTY,\DTZ)$, where $\DX$, $\DY$ and $\DZ$ are the desired corrections needed for the EEF in the Cartesian space w.r.t. the current location. 

\subsection{Multiview Multimodal Representation Learning}

\begin{wrapfigure}{r}{0.22\textwidth}
  \begin{center}
    \includegraphics[width=0.45\columnwidth]{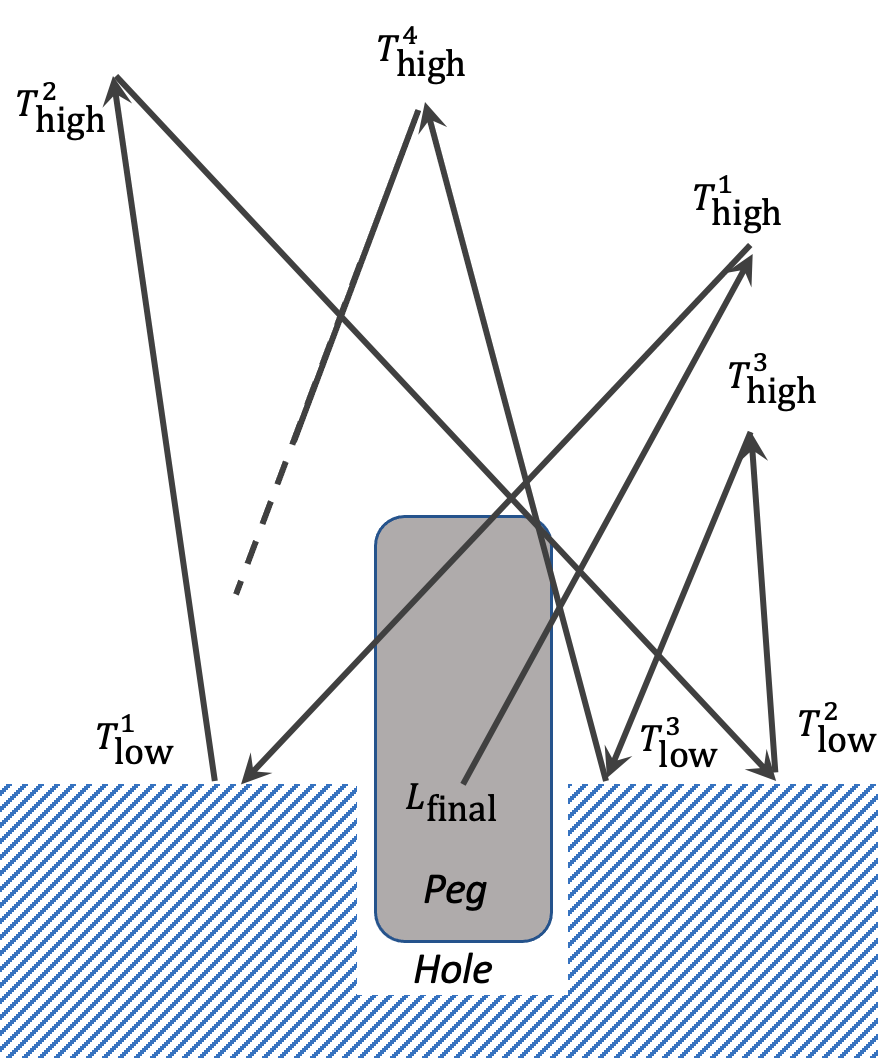}
  \end{center}
  \caption{\label{fig:random_data_collection} Illustration of the longitudinal section of the random data collection applied in Algorithm \ref{Algo:backward} around the peg. Note that it switches between low and high targets and that it randomly fills the space around $L_\textrm{final}$.}
\end{wrapfigure}

Combining vision and force to tackle general insertion tasks is necessary because the force supplies information about the area close to the hole \cite{inoue2017deep}, e.g., in cases of occlusion, while vision provides information when there is no overlap between the peg and the hole. In our previous work, we showed that a wrist-mounted camera offers a notable advantage over fixed cameras when it comes to generalization, as it can exploit problem symmetries and avoid occlusions. However, we had to add the requirement that the robot touch the socket in order to avoid the ambiguity problem, since we aim to evaluate the distance between two points in world coordinates using a single image. In the present scheme, we used two symmetric cameras, which enabled us to bypass the one image ambiguity problem and extract depth information while avoiding occlusion during the entire insertion trajectory (if one camera's view is occluded, the other one has a clear view). After extensive evaluation, we decided to position each camera at a 45-degree angle with respect to its respective finger opening, resulting in a good view of the scene as well as of the object between the fingers.

\subsection{Data Collection and General Backward Learning \label{secsub:backward_learning}}

The two-camera scheme theoretically allows us to recover the distance between two points, avoiding the vision ambiguity problem we had in previous work \cite{spector2021insertionnet} (see Figure \ref{fig:image_force_network}). General backward learning leverages the two cameras to collect data not only after touching the hole's surface but also along the moving trajectory. Similar to InsertionNet, the robotic arm is positioned in its final position, $L_{\textrm{final}}$. Then, for each iteration, we sample two points from a normal distribution: one is $T_\textrm{high}$, which is positioned in a random location above the hole, and the second one, $T_\textrm{low}$, is randomly positioned around the hole's height\footnote{We tested also a uniform distribution instead of a normal one, but it resulted in a large fraction of the points being located far from the hole, indicating a less effective training scheme.}. 
The correction is defined by $\DELTA_{\textrm{correction}}\triangleq L_{\textrm{final}} - T_{\textrm{random}}$, where $T_{\textrm{random}}$ is a high or low point. We built such a data set, denoted by $D$, based on the set of training tasks $\tau\triangleq\{\tau_i\}_{i=1}^{N_\tau}$, where for each task, $\tau_i$, there is a corresponding $L_\textrm{final}^i$. For each task, a randomized set of points $\{T_{\tau_i, j}\}_{j=1}^{N_{\tau_i}}$ is generated, yielding starting and end points for tasks and the initial random points $D\triangleq \cup_{\tau_i \in \tau} \{ D_{\tau_i , j} \}_{j=1}^{N_{\tau_i}}$ and $D_{\tau_i , j} \triangleq [L_\textrm{final}^i, T_{\tau_i, j}]$. For each $D_{\tau_i , j}$, we define a corresponding correction, $\DELTA_{\tau_i, j}\triangleq L_\textrm{final}^i - T_{\tau_i, j}$. This procedure is presented in detail in Algorithm \ref{Algo:backward}.


{
\SetAlgoNoLine
\begin{algorithm}[!t]
\caption{Data Collection and General Backward Learning for Task $\tau$}\label{Algo:backward}
    \DontPrintSemicolon
    \SetKwFunction{Fupdate}{update\_tree}
    \SetKwFunction{Ftop}{top\_actions}
    \KwIn{Maximum Randomized Location $r_0=10[\textrm{mm}]$; Maximum Randomized Angle $\theta_0=10^\circ$; Maximum Height $z_{\textrm{max}}=50[\textrm{mm}]$; Force threshold $F_{\textrm{th}}$; Momentum threshold $M_{\textrm{th}}$; Points record $P_{\textrm{rc}}$ ; Reaching time = $t_\textrm{max}$, number of data points $N_\tau$, Buffer of Data $D$}; 
    \textbf{Init:} Insert plug into the hole, register final pose $L_\textrm{final}$ \\
    \textbf{Set:} $(\overbrace{x_0,y_0}^{X_0},z_0, \overbrace{\theta_{x_0}, \theta_{y_0}, \theta_{z_0}}^{\Theta_0})=L_\textrm{final}$ \\
    \For{$i \leftarrow 1,\ldots,N_\tau$}
    {
        $p_X^0,p_X^1 \sim \mathcal{N}(X_0,r_0)$\\
        $p_\Theta^0,p_\Theta^1\sim \mathcal{N}(\Theta_0,\theta_0)$  \\
        $z^1 \sim  \textrm{Uniform}[z_0, z_{\textrm{max}}]\quad \textrm{random height above $z_0$}$ \\
        Set target $T^i_\textrm{high}=(p_X^0,z_1, P_\Theta^0)$ \\
        Set target $T^i_\textrm{low}=(p_X^1,z_0, P_\Theta^1)$ \\
        \For{$j \leftarrow [\textrm{high},\textrm{low}]$}{        
        Move Robot to $T^i_j$ using an impedance controller\\ 
        \While{Robot did not reach $T^i_j$} {
            \If{time \textup{mod} $t_\textrm{max}/P_{\textrm{rc}}$ } {
                \textrm{Register Pose} $E = \{x , y , z, \theta_{x}, \theta_{y}, \theta_{x}\}$ \\
                Capture two $45^\circ$ cameras into $\IMG$\\
                Capture force $F$ and momentum $M$ \\
                \textrm{Set} $\textrm{Obs} = \{ \IMG, F, M \}$ \\
                \textrm{Set} $\textrm{Action}_i = \{L_\textrm{final}-E\}$ \\
                Push $\{\textrm{Obs}, \textrm{Action}\}$ into $D$ \\

           }        
        }
    }
    \Return $D$ \\
    }
\end{algorithm}
}

\subsection{Multi-Objective Regression Residual Policy Formulation\label{subsec:MultiObjective}}
\begin{figure}[h]
\centering
\includegraphics[width=0.48\textwidth]{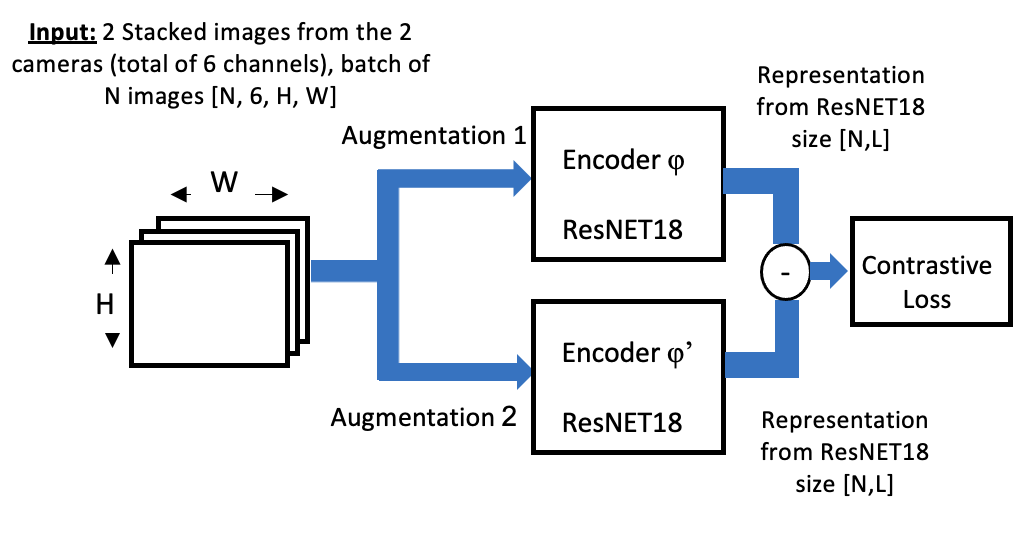}
\caption{\label{fig:contrastive_network} Architecture of the contrastive system.}
\end{figure}

\begin{figure}[h]
\centering
\includegraphics[width=0.48\textwidth]{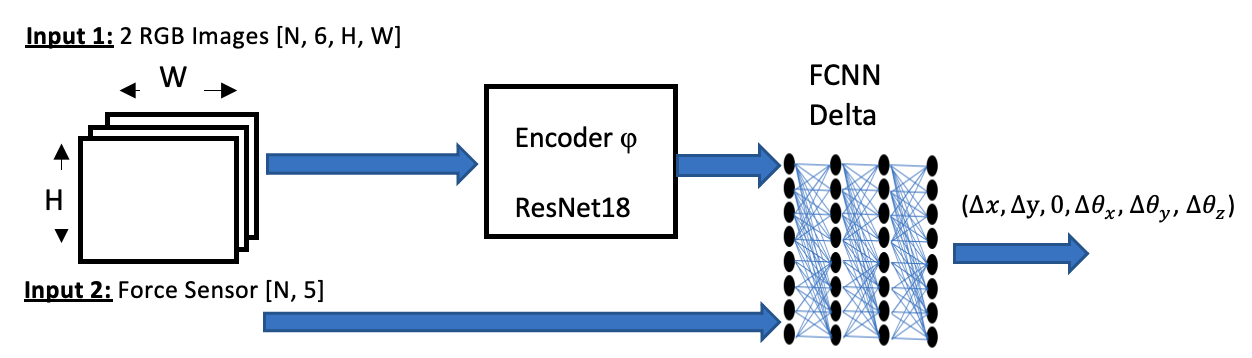}
\caption{\label{fig:image_force_network} Delta Architecture for the Delta policy.}
\end{figure}

\begin{figure}[h]
\centering
\includegraphics[width=0.48\textwidth]{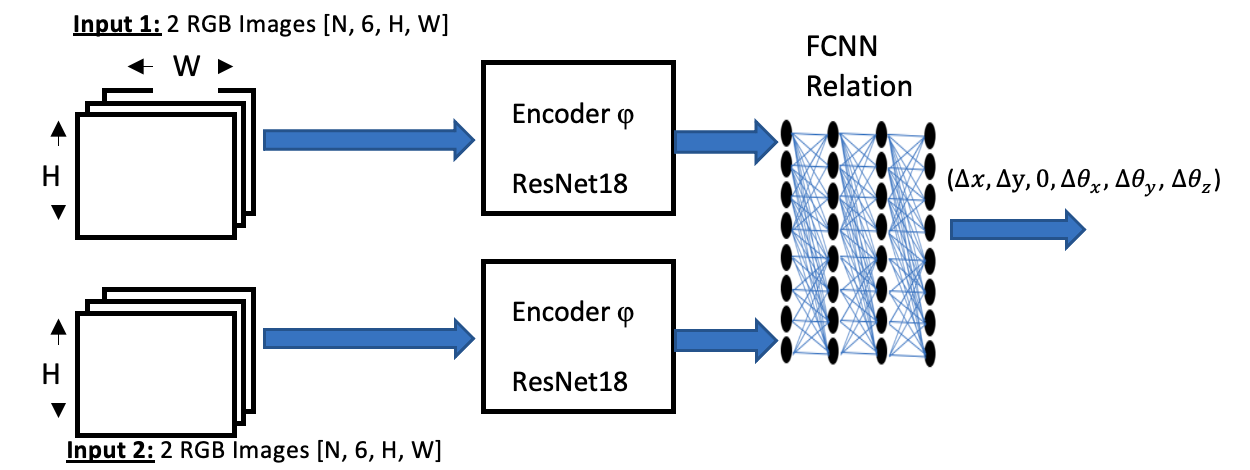}
\caption{\label{fig:double_image_network} Relation Architecture based on a dual-input image for the Relation policy.}
\end{figure}

At the heart of the proposed methodology is a ResNet18 network \cite{he2016deep_resnet}, denoted by $\varphi$. In the following, we enlist $3$ architectures and use $\varphi$ to synergize the network's performance. We  note  that the  three architectures presented below (Figures \ref{fig:contrastive_network}, \ref{fig:image_force_network} and \ref{fig:double_image_network}) share a common network that is simultaneously trained on all 3 architectures, i.e., $loss=l_\textrm{contrastive} + l_\textrm{Delta}+ l_\textrm{Relation}$.

\noindent \textbf{(1) Contrastive Loss and Contrastive Architectures:} We trained $\varphi$ using contrastive techniques \cite{chen2020simple_contrastive}, where it learns relevant features for the task at hand without any specific labels (Figure \ref{fig:contrastive_network}). It is based on the InfoNCE loss \cite{oord2018representation} and is used similarly to \cite{laskin2020curl}.
Specifically, by  stacking two images from the two cameras, we implicitly obtain a depth registration of the plug and socket. We use this information to augment in different ways the stacked image and transfer it to $\varphi$ and $\varphi'$, the latter being a network with the same weights that is updated using a Polyak averaging $\varphi' = m \varphi' +(1-m)\varphi$ with $m=0.999$. As we show hereunder, the contrastive loss shrinks our solution's variance, thereby creating a more robust policy. Also, the contrastive loss enables the use of stronger augmentations in the training. 

\noindent \textbf{(2) Delta Movement Loss and Architecture:} Network $\varphi$ is trained to predict the $\DELTA_{\textrm{correction}}$ (Section \ref{secsub:backward_learning}) using $O$ (relevant collected images, forces, and moments).

\noindent \textbf{(3) Relation between Loss and Architecture:} Based on previous concepts from vision (e.g., RelationNet \cite{sung2018learning}), we introduce a relation loss in order to facilitate one-shot learning \cite{sung2018learning}, as well as enable multi-step insertion and improve the exploitation of the collected data. When training with this loss, we use the same sample set, $D$, to calculate the delta movement between two images of the same plugging task, $\IMG_i$ and $\IMG_j$, $j\ne k$. The ground truth in this case is calculated by the difference $\DELTA_{\tau_k, j}=\DELTA_{\tau_i, j}-\DELTA_{\tau_i, k}$.
Of note, the augmentations of $\IMG_i$ and $\IMG_j$ need to be consistent.

\subsection{Main Policy and Residual Policy}
The proposed algorithm uses a policy that is composed of two policies: (1) A Main Policy, $\pi_\textrm{Main}$, which approximates the location of the hole (see \cite{spector2021insertionnet}); (2) A Residual Policy, $\pi_\textrm{Residual}$, which is activated at distance $R$ from the hole surface and does the actual insertion. We note that the $\pi_\textrm{Residual}$ action is of the form $\DELTA=(\DX,\DY,0,\DTX,\DTY,\DTZ)$, where the $\DZ$ is controlled independently (see Algorithm \ref{Algo:Residual}), since the table's height is obtained from prior knowledge or from the depth camera.

\subsection{Inference and Delta Policy vs. Relation Policy}

We introduce two policy architectures,  each one with its distinct pros and cons. On the one hand, the Delta Policy has a similar structure to a ResNet model, which is known for its accuracy in tasks like ImageNet but is limited in learning from only a few samples. On the other hand, measuring similarities tends to be less accurate and robust but offers a large advantage when tackling few-sample problems. When applying the trained network $\varphi$ at the inference time, we use either the Delta architecture or the Relation architecture. The choice depends on the use-case. For one-shot or multi-step insertion tasks, we use the Relation architecture, since it can generalize better in these tasks. For all other tasks, we use the Delta architecture.

\subsection{Data Augmentation}
\label{subsec:augmentation}
In order to facilitate robustness as well as generalization over color and shape, we used various augmentations. We discovered that the order as well as the properties of each augmentation has a large effect on generalization.
\noindent \textbf{Visual Augmentation:}
For Delta loss /Relation loss, we used "resize, random crop, color jitter, translation, rotation, erase, random convolution". For the contrastive loss, we used "RandomResizedCrop, strong translation, strong rotation, erase". We note that all the samples in a batch have similar augmentations.

\noindent \textbf{Force Augmentation:} 
In insertion tasks, the direction of vectors (F, M), not their magnitude, is the most valuable factor. Therefore, the input to the NN is the direction of the force and moment vectors. 
\section{EXPERIMENTS} \label{sec:experiments}
In this section, we describe the experiments conducted to validate our methodology, in which we address the following questions:
\begin{enumerate}[wide, labelwidth=!, labelindent=0pt]
    \item Scalability - How does our algorithm scale to various insertion tasks in terms of accuracy and time?
    \item Robustness - How robust is our algorithm to  uncertainties in illumination, grasping, and visual localization?
    \item Transferability - How well does the trained policy transfer to other tasks?
    \item Multi-step insertion - How well does the relation method learn $3$-step insertion tasks?

\end{enumerate}

\begin{figure*}[ht]
\centering 
\includegraphics[width=0.99\textwidth]{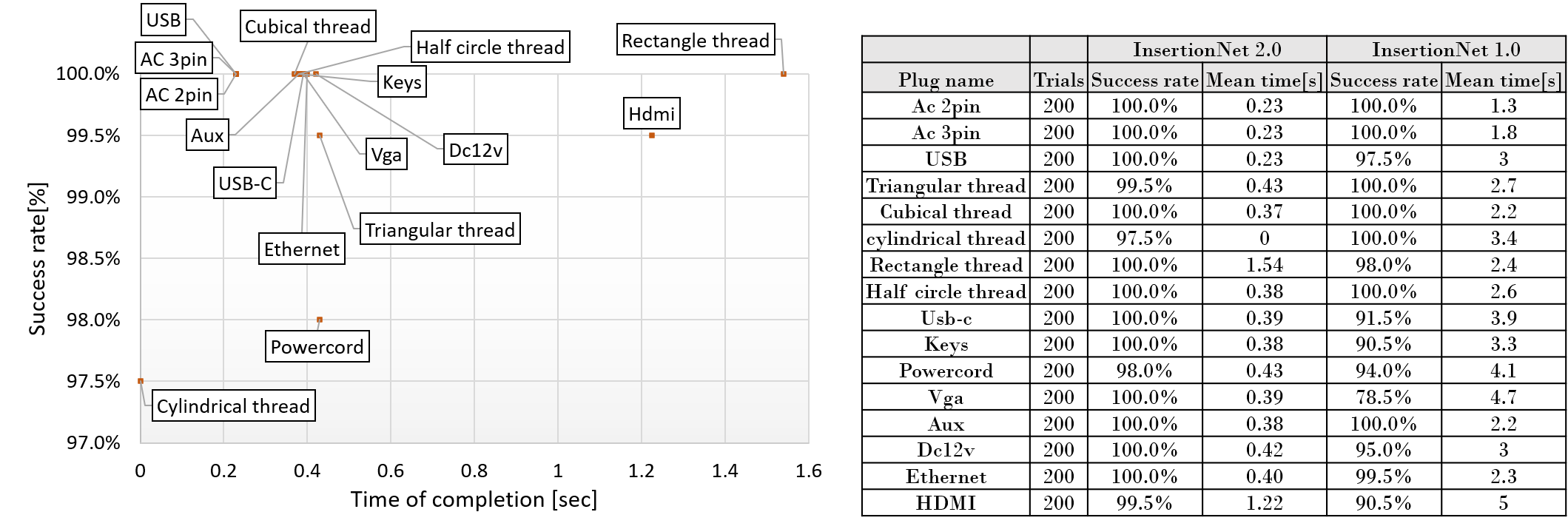}
\caption{Graph (left) presenting the overall performance on the 16 insertion and threading tasks averaged over 200 trials.  Table (right) present the overall performance compared to inesrtionNet 1.0\cite{spector2021insertionnet}.
\label{fig:all_results}}
\end{figure*}

{
\SetAlgoNoLine
\begin{algorithm}[b]
\caption{\strut Residual Policy}\label{Algo:Residual}
    \DontPrintSemicolon
    \KwIn{Target Point $T$; Transition Radius $R$; Final position image $\IMG_F$; Lowering rate $\Delta_Z$; Final height in $z$-axis $z_\textrm{final}$; 
    Choose either $\pi_\textrm{Residual}$ = $\pi_\textrm{Delta}$ (Fig. \ref{fig:image_force_network}) or $\pi_\textrm{Relation}$ (Fig. \ref{fig:double_image_network}) according to task in hand.
    }
    \textbf{Init:} $NextPoint=T$, $\textrm{FinishFlag} =False$\\
     
     \While{not $\textup{FinishFlag}$} {
            Robot move to $NextPoint$ \\
            $ CurrentPoint =$ Measure $(x,y,z,\theta^x, \theta^y, \theta^z)$ \\ 
            \If{$dist(CurrentPoint,T) < R$} {
                
                $\Delta=\pi_{Residual}(O, \IMG_F) $ \\
                $\delta_z= (0,0,R/\Delta_Z$,0,0,0)\\
                $ NextPoint= CurrentPoint+\Delta+\delta_z$ \\
            
                \If{$z < z_\textrm{final}$} {
                    $\textrm{FinishFlag} = True$\\
                }
           }        
        }
\end{algorithm}
}

\subsection{Experimental Setup}

\textbf{Data Collection and Training:} We start with  collecting 150 data points using the General Backward Learning approach (Algorithm \ref{Algo:backward}). In the second stage, we train a residual policy neural net architecture for each plug (described in Section \ref{subsec:MultiObjective}) with the proposed visual augmentations of Section \ref{subsec:augmentation}. Based on 150 samples, we generate 192,000 training samples\footnote{We note that it includes the augmentations as well as pairs of samples for the Relation policy} (until convergence is reached, at around 1500 repeats on 64 sample-batches), using a 2 GPU GeForce RTX 2080 Ti. The total training time takes $5-15$ minutes.

\textbf{Testing Procedure:} In each experiment, we examine one plug-socket setup. We generate test points, denoted by $N_{\textrm{test}}$, around the testing socket in different locations above the hole, by uniformly sampling $|N_{\textrm{test}}|$  target point $T$ from $B_{xy} \times B_\theta \times  B_Z $.  We note that $T$ is the localization point and that it contains errors from grasping, misalignment and localization. Afterward, the robot sends $T$ to the process described in Algorithm \ref{Algo:Residual}. 
A trial is considered successful if the robot inserts the peg/plug into its designated hole with tolerance $z-z_0<1[\textrm{mm}]$. An unsuccessful trial is when the duration it takes to complete the task exceeds $10[\textrm{sec}]$. We repeat each insertion trial $200$ times (with random initialization) in order to achieve statistically significant results. In order to examine realistic scenarios, the pegs are not fixed to the gripper (slippage between the peg and fingers is possible) during insertion, and the 200 trials are consecutively applied in a row to intentionally create misalignment during the insertions. This solution does not require contact with the hole surface to correct the movement, and the execution speed solely depends on the robot's controller. Therefore, we subtract the mean time from the reaching time to better reflect the difficulty in solving an insertion. The difference in completion time between the different tasks is due to the small clearance and soft touch needed to insert the objects (impedance controller + algorithm corrections).

\subsection{Scalability for Different Insertion Setups}
\label{sec:scalability}
We tested our Delta policy on the same 16 real-life insertion and threading tasks we used to test our previous InsertionNet version \cite{spector2021insertionnet}: AC 2-pin, AC 3-pin, Aux, DC12V, ethernet RJ45, HDMI, keys, power cord, USB, USB-C, VGA, HDMI, AUX, circle "pulley", half circle "pulley", triangle "pulley", and cubical "pulley". 
Figure \ref{fig:all_results} summarizes the success rate of and time needed to complete each of the 16 different insertion cases, averaged over 200 trials. The lowest score was 97.5$\%$, in the cylindrical thread, while the longest mean time was 1.54[sec], in the rectangle thread. Using the current algorithm, we improved our performance in the VGA task from $78.5\%$ to 100$\%$, and from 90.5$\%$ to 99.5$\%$ on HDMI plugs. In general, this algorithm achieved a near-prefect score on all 16 insertion tasks. 

The improvement in the VGA and HDMI tasks can be attributed to 3 factors: 1. Minimal contact motion helped us avoid friction issues. 2. The two-camera scheme increased the accuracy, because there
is no concealment at all times. 3. The ur5e controller seems to be more accurate than the Franka Emika Panda robot.

\subsection{Sample Efficiency}
We assessed how many samples are required to successfully accomplish an insertion task. In order to calculate the relation between the number of samples and the success rate or the required time for completion, we trained $8$ different policies using $10$, $15$, $20$, $25$, $30$, $50$, $100$ and $200$ samples. Each policy was tested $50$ times on the AC 2-pin plug task. The success rate and time for completion vs. samples required are depicted in Figure \ref{fig:number_of_samples}. The  success rate rapidly increases between $0$ and $40$ samples, reaching 100$\%$ with 40 data points, while the time for completion steeply declines between $0$ and $150$ samples, reaching nearly $0[\textrm{sec}]$ at $150$ samples. Based on these results, we can reverify that $150$ samples are sufficient for the  Section \ref{sec:scalability} experiments.

\subsection{Utilizing Unlabeled Data and Contrastive Loss Impact}
\textbf{Utilizing unlabeled data}: We evaluated whether incorporating unlabeled data improved the performance on insertion tasks. We collected 20 labeled data on AC 2-pin plugs and 150 images of unlabeled data, acquired by moving the robot around the designated hole with the UR5E pad. Utilizing unlabeled data in our experiment improved the Relation Policy results from 20$\%$ to 100$\%$, and the Delta policy results from 70$\%$ to 100$\%$. In addition, the movement for both policies is more stable, with less variance in the movement. \\
\textbf{Contrastive loss impact}: We evaluated our policies' behavior learned with contrastive and without contrastive learning on 3 tasks: AC 2-pin, USB and yellow thread. While we did not find any noticeable difference in the Delta Policy's performance, the Relation Policy is more stable with the contrastive loss addition. The stable behavior is reflected in an approximately 20$\%$ higher success rate with contrastive loss addition and in a smaller variance in the completion time. This gap was mostly significant when small data sets were used. This result is consistent with the literature, since contrastive loss and triplet loss have similar goals and mechanism that were shown to be useful in learning embedding for one-shot learning in vision \cite{hoffer2015deep}.

\subsection{Robustness}
\begin{figure}[h]
\centering
\includegraphics[width=0.34\textwidth]{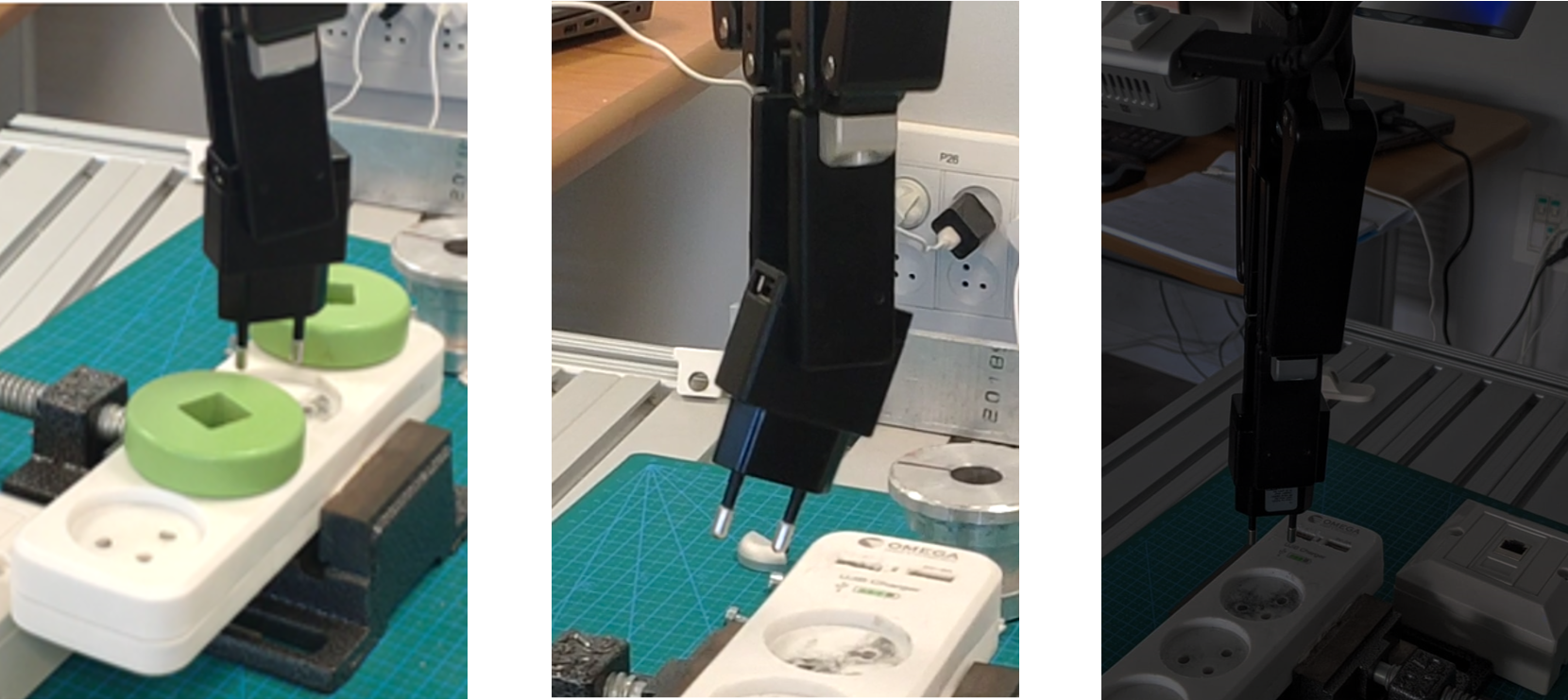}
\caption{\label{fig:robustness} Testing our method’s robustness to obstacles (left), strong grasping errors (middle), and illumination (right).}
\end{figure}

Another aspect of interest is \textbf{Delta Policy's} robustness to changes (see Fig. \ref{fig:robustness}) in grasping up to 45 degrees, strong illumination changes (i.e. turning the room lights on and off), placing obstacles randomly around the insertion hole, and changing the board's location/orientation during movement. We first collected 300 data points, 100 with a grasping of 0 degrees, 100 with 15 degrees, and 100 with -15 degrees, in order to learn a policy robust to strong changes in grasping. Our results (see attached video) show that our method overcame these challenges obtaining a perfect score of 50/50 while exhibiting a minor decline in the execution time in comparison to the basic plug. Note that without sampling data with a grasping orientation, our model does not generalize well to large angular errors. With regard to the \textbf{Relation Policy}, while the Delta Policy exhibits strong robustness to obstacles, the relation network struggled to accomplish the task without collecting addition data with obstacles. In addition, we observed that the relation policy needs more data in order to generalize to strong changes in grasping.  

\begin{figure}[b]
\centering
\includegraphics[width=0.45\textwidth]{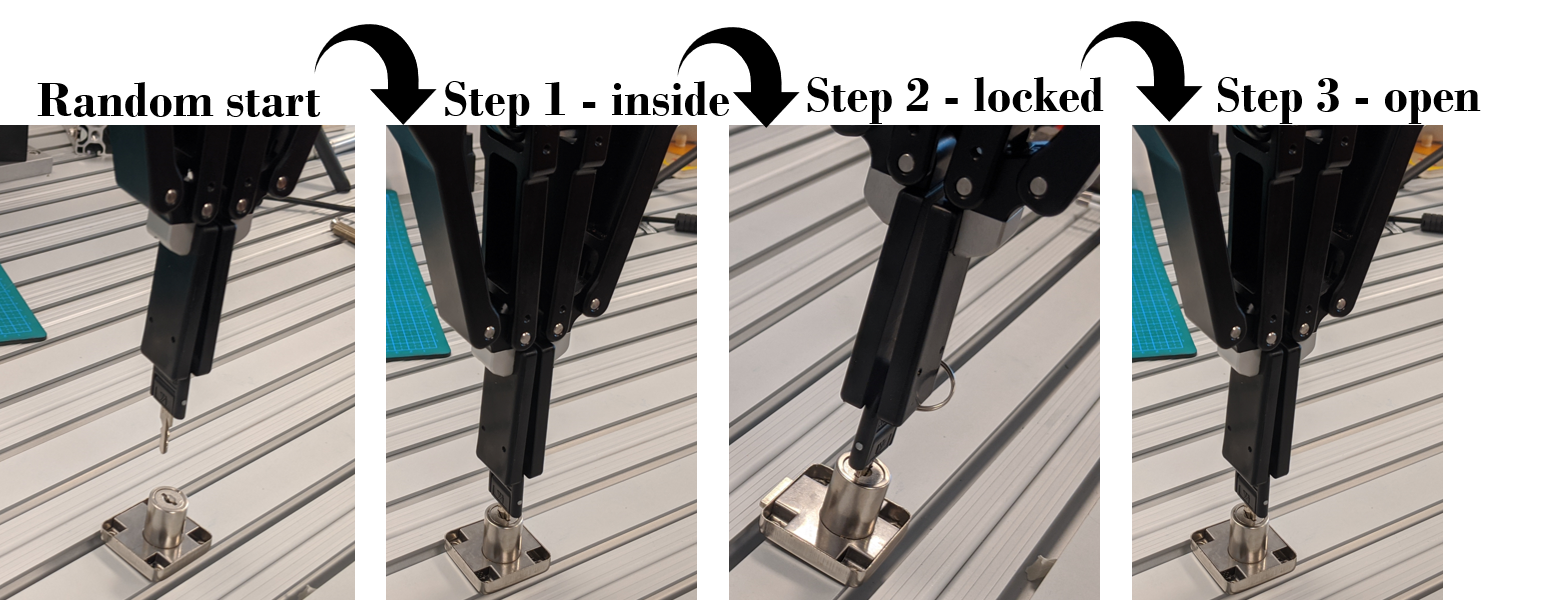}
\caption{\label{fig:multi_step} Multi-steps insertion task. }
\end{figure}

\subsection{Transferability}
\begin{figure}[h]
\centering
\includegraphics[width=0.42\textwidth]{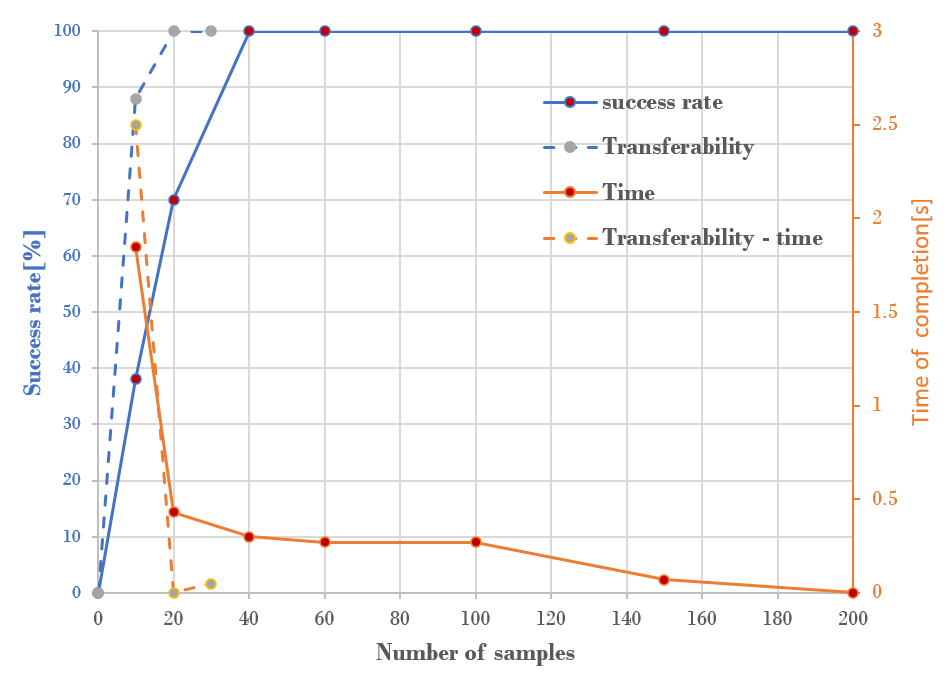}
\caption{\label{fig:number_of_samples} The success rate (blue, lines) and time for completion
(orange, lines) with (dotted line) and without (full line) a pre-trained
policy.}
\end{figure}

We also evaluated how well a trained policy is transferred to other tasks. The policy was trained on all the 15 plugs except the red, half-circle thread. \\
\textbf{One-shot learning with the Relation Policy:}
We initially took an image of the final state, i.e., when the plug is inside the hole, and tested our pretrained policy on an unseen plug that is somewhat similar to the learned data (the red, half-circle thread). Our Relation Policy achieved perfect results, 50/50, on the red, half-circle thread, with no decline in the duration time relative to previous results. We note that the Delta Policy on the same task scored 0/50 without any additional data.\\
\textbf{Delta Movement Policy:}
We measured how much additional data is needed in order to preform well on the red, half-circle thread task. As shown in Fig. \ref{fig:number_of_samples}, only 20 samples are needed to archive good results when dealing with unseen objects. This is almost 10$\%$ of the original amount needed (150 samples), demonstrating that our algorithm can share information between different insertion tasks.

\subsection{Multi-Step Insertion}
Lastly, we evaluated our system's ability to tackle multi-insertion tasks. Since in multi-insertion tasks, like locking a door, it is harder  to  collect  data  and  verify that each step can be completed, we used our Relation Policy. The problem we tried to address, presented in Fig. \ref{fig:multi_step}, is composed of 3 steps: inserting the key, turning the lock, and then turning back. For each step, we pre-saved an image of the state with both of our 45-degree cameras. The execution followed Algorithm \ref{Algo:Residual} with the Relation Policy and our similarity function to switch between steps. Although our algorithm used the regular backward data collection Algorithm \ref{Algo:backward} and did not visit the locking and insertion states (visiting states only above or touching the hole surface), we were able to accomplish the task, scoring 50/50.

\section{DISCUSSION and FUTURE WORK} \label{sec:discussion}

 
The presented version of InsertionNet successfully achieved a greater than $97\%$ success rate over 16 different tasks while achieving an execution time limited only by the robot hardware, i.e, the controller's speed. In future work, we would like to reduce the execution time further by incorporating the same algorithm on a parallel manipulator, known for its speed. In addition, we present a relation network scheme that facilitates one-shot learning and solves multi-step insertion tasks. We intend to examine this method's ability to tackle other, more complex multi-step tasks. Finally, we find that unlabelled data can be utilized using the contrastive loss scheme to minimize sample complexity, specifically, for the Relation Policy. Further exploration of the benefits and limitation of unlabeled data is left for future work.

\bibliographystyle{IEEEtran}
\bibliography{ICRA.bib}

\end{document}